# Theory Refinement on Bayesian Networks

Wray Buntine
RIACS and AI Research Branch
NASA Ames Research Center, Mail Stop 244-17
Moffet Field, CA 94035, USA
Phone: +1 (415) 604-3389
wray@ptolemy.arc.nasa.gov

## Abstract

Theory refinement is the task of updating a domain theory in the light of new cases, to be done automatically or with some expert assistance. The problem of theory refinement under uncertainty is reviewed here in the context of Bayesian statistics, a theory of belief revision. The problem is reduced to an incremental learning task as follows: the learning system is initially primed with a partial theory supplied by a domain expert, and thereafter maintains its own internal representation of alternative theories which is able to be interrogated by the domain expert and able to be incrementally refined from data. Algorithms for refinement of Bayesian networks are presented to illustrate what is meant by "partial theory", "alternative theory representation", etc. The algorithms are an incremental variant of batch learning algorithms from the literature so can work well in batch and incremental mode.

## 1 Introduction

Theory refinement is the task of updating a domain theory in the light of new cases. The key idea is to use the expert's prior domain knowledge to prime a learning system during the knowledge acquisition process. Subsequent refinement of theory proceeds by having the learning system accept examples or ask key questions of the expert. Shapiro (Shapiro, 1983), for instance, developed a comprehensive theory and suite of algorithms for the task of refining Horn clause theories (logic programs). Ginsberg et al. applied a more heuristic approach to the refinement of a rule base in the context of medical diagnosis (Ginsberg et al., 1988). Recent research in this area (Ourston and Mooney, 1990; Towell et al., 1990) grew out the need to make the many inductive learning algorithms available more knowledge intensive, so they can mimic some of the perceived benefits of analytic learning methods such as explanation based learning. But this research faces the problems of "imperfect and uncertain domain theories" and "noisy training cases" not well handled by analytic methods.

A recent example of this hybrid learning approach is as follows (Towell et al., 1990): a rule-base of knowledge about the domain is transcribed into a neural network to initialize the network; the new training cases are then run in a back-propagation algorithm to refine the network. This approach addresses the following research question: how can we build a learning algorithm that covers the full spectrum from theory refinement, to standard batch learning (starting with a non-informative theory, and assuming learning occurs from just one batch of cases), to incremental learning (assuming new cases come in smaller batches and the theory is gradually refined)?

A second recent example of theory refinement is of Bayesian networks sometimes used in medical expert systems (Lauritzen and Spiegelhalter, 1988). While experts can set up an appropriate graphical structure and estimate the needed probabilities, new examples may arrive on a daily basis so the expert system needs to be refined. Spiegelhalter et al. argue that the expert's experience and confidence in setting up the initial model needs to be quantified (Spiegelhalter and Lauritzen, 1989) (for instance, how many examples was it based on) in order to do refinement carefully. It could be that the expert's initial model is based on many cases and is very reliable, and the 10 new noisy cases obtained happen to be unusual so they wrongly suggest the expert's initial model requires major refinement. Spiegelhalter et al.'s approach addresses a second research question: given some new and possibly anomalous cases, when do we start refining, how drastically do we refine, and when do we disregard the anomalous cases as noise? Spiegelhalter, however, did not address the issue of refining the structure of a Bayesian network, only the continuous parameters of the probability distributions.

This paper considers these two broad research questions together. The approach to theory refinement suggested is as follows: the learning system is primed with a partial theory supplied by a domain expert, and thereafter maintains its own alternative theory representation which is able to be interrogated by the domain expert and able to be incrementally refined from data. Furthermore, the partial theory is such that it can initially be null, and that it incorporates a quantification of the expert's experience so that the "right" amount of refinement is done given new cases. Another approach to learning networks that incorporates a partial theory is given by by (Srinivas et al., 1990).

The general approach developed here is based on Bayesian principles for belief updating that form the basis of several learning algorithms (Buntine, 1990b; Cooper and Herskovits, 1991). The principles specify precisely a "normative" approach to theory refinement, and the approach suggested here approximates this. The normative property is a claim that the principles set a standard which other theory refinement or learning algorithms must approximate; if they fail to do so they will return poorer refined/learned theories on average. Another popular learning framework in the computing area is uniform convergence, of which the PAC model is an instance (Haussler, 1991). This is an approach that approximates the normative Bayesian approach when sample sizes are large. Several researchers have reported (unsurprisingly) that the Bayesian approach is superior with smaller size training samples (Buntine, 1990b; Opper and Haussler, 1991) in a range of batch learning problems.

Some previous methods for learning Bayesian networks (Geiger et al., 1990; Spirtes and Glymour, 1990; Verma and Pearl, 1990; Srinivas et al., 1990) are closer to the large sample uniform convergence framework because they assume independence information can be unambiguously determined. Some of these algorithms also make the assumption (Geiger et al., 1990; Spirtes and Glymour, 1990; Verma and Pearl, 1990) that the unknown probability distribution is a DAG-isomorph (Pearl, 1988). This means all independencies in the problem must be perfectly captured by some Bayesian network, which may not be the case in a particular problem (for instance, all non-chordal Markov networks are not DAG-isomorphic). These algorithms can seemingly "discover causality from data", but existence of some "causality" is immediate from the assumption of DAG-isomorphism. How restrictive will this assumption be in practice and how sensitive are the algorithms to its failure? The approach here in contrast requires that some ordering (possibly causal) of the variables is supplied to the system. This assumes nothing about the underlying probability distribution because a Bayesian network can always be found for some ordering. The algorithms presented do, however, assume that every example in the training sample has variable values fully specified. (While this assumption can be relaxed, it can involve considerable computational cost if done properly.)

In the approach presented here, the initial partial theory obtained from the domain expert is interpreted as *a prior* information about the space of possible theories, and the alternative theory representation is interpreted as a subspace of alternative theories that are reasonable *a posterior*, represented in a compact form. Simple learning approaches approximate this space of alternative theories by taking a single high posterior structure (Cooper and Herskovits, 1991; Buntine, 1990a) however experiments show that averaging over a larger sized space yields considerable improvement (Buntine, 1990a)[1]. This improved performance corresponds to the improved accuracy gained in the $TOPN$ system when the system approximates posteriors using a thousand alternative disease sets instead of a single disease set (Henrion, 1990).

A space of alternative theories is difficult to present to a domain expert but can be readily summarized in several ways for expert interrogation during theory refinement: two approaches are described here. The theory refinement algorithm of course applies Bayes theorem to this space of alternatives. To generate a space of reasonable alternatives, it does a search of the space of high posteriors in a similar style and with the same motivation as the $TOPN$ system and the Bayesian averaging method for trees (Buntine, 1990a).

The theory refinement approach is developed here for Bayesian networks. These networks are first introduced and then the representation of partial theories and their transformation to a prior is described. The representation for alternative theories is described, and then the theory refinement and interrogation algorithms are presented. These major sections describe the approach but assume that conditional probability distributions for each node in a Bayesian network are represented with a full conditional joint distribution, and that all values of variables are supplied with each training case. Of course, in larger practical systems, these two assumptions rarely apply. The final section describes how noisy-or gates and other lower-dimensional conditional distributions can have their parameters learnt within the same theory refinement framework.

## 2   Bayesian Networks

Bayesian networks specify dependence properties between variables by using a directed acyclic graph. They describe probabilistic models useful for non-directed classification. That is, one can predict (and compute likelihoods for) one subset of variables from any other. In contrast, class probability trees (Quin-lan, 1986; Buntine, 1990a) *only allow directed classification* because they only yield predictions about a special target variable usually referred to as the class.

---

[1]Similar results are reported in (Spirtes et al., 1990), although their justification is different.

Figure 1 shows a simple Bayesian network. The set of variables that have outgoing arcs to a variable $x$ are called the parents of the variable $x$. Each variable also has an associated conditional probability table which gives probabilities for different values of the variable conditioned on values of its parent variables. For instance for the graph in the figure, we need values for $Pr(e|a)$, $Pr(d|a,b)$, $Pr(g|c,d)$, etc., because $a$ is the only parent of $e$, etc. Given the *parent structure* specifying the network and the *conditional probability tables*, methods exist for computing arbitrary conditional and marginal likelihoods between variables (Lauritzen and Spiegelhalter, 1988).

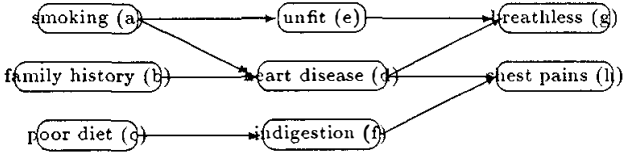

Figure 1: Bayesian network for a simple system

The following notation is used here. A Bayesian network consists of a set of discrete variables $\mathcal{X}$ where each variable $x \in \mathcal{X}$ has a set of parent variables $\Pi_x$. The full parent structure is denoted $\Pi$. For instance, for the graph in the figure, $\Pi_e = \{a\}$, $\Pi_d = \{a,b\}$, etc. The set of possible values for the variable $x$ is $v(x)$ and for the cartesian product of variables in $\Pi_x$ is $v(\Pi_x)$. For instance, if $a$, $b$ and $d$ are boolean, then $v(a) = \{true, false\}$, and $v(\Pi_d) = \{(true, true), (true, false), (false, true), (false, false)\}$. Also, $m_x$ denotes the cardinality of $v(x)$.

Given an assignment $I$ to the variables in $\mathcal{X}$, $\mathcal{X} = I$, denote corresponding assignments to $x \in \mathcal{X}$ by $I_{|x}$, and to $\Pi_x \subset \mathcal{X}$ by $I_{|\Pi_x}$. For instance, if $\mathcal{X} = \{u, v, w\}$ and $\Pi_u = \{v, w\}$ for $u$, $v$ and $w$ boolean, then if $I = (true, false, true)$, then $I_{|u} = true$ and $I_{|\Pi_u} = (false, true)$. Also, $\theta$ denotes the matrix of conditional probabilities for $x$ given that the parent variables are $\Pi_x$ and conditioned on their values. So $Pr(x = i \mid \Pi_x = j, \Pi_x, \theta) = \theta_{x=i|j}$. With these, we are able to determine the probability of the full set of variables $\mathcal{X}$ using the standard expansion

$$Pr(\mathcal{X} = I \mid \Pi, \theta) = \prod_{x \in \mathcal{X}} \theta_{x=I_{|x}|I_{|\Pi_x}} .$$

This gives the likelihood for a single example given $\Pi$ and $\theta$, and a product of these forms gives the likelihood for an independently and identically distributed training sample, used in calculating various posteriors.

## 3  Partial Bayesian networks

An initial partial theory given by the expert is to be transformed to a prior probability over the space of theories. Since a Bayesian network is fully specified by a parent structure $\Pi$ together with conditional probabilities $\theta$, an initial partial theory then somehow specifies a prior distribution $Pr(\Pi, \theta)$. This section describes the information obtained from the expert and how it is converted into a prior on Bayesian networks.

Experience shows that experts are often able to suggest roughly which variables influence which. This is because experts are usually better at expressing qualitative knowledge than quantitative, and because weak domain theories often indicate influence but not its exact equational form. If variables are ordered according to time of occurrence, for instance family history of heart disease pre-dates heart disease, then many of the potential influences (those following in time) are made impossible. The partial theory obtained from the expert is an ordering of variables and a Bayesian network specified pictorially in shades of grey. Black arcs indicate definite parents (with a prior of 1). Missing arcs indicate definite non-parents (with a prior of 0). Grey arcs indicate parents whose status we are uncertain about, with prior belief proportional to the grey level (or to allow greater range, with log prior mapped to the grey level). This tells the theory refinement algorithm how eager it should be to modify a potential parent's status in the light of new evidence.

We ask the expert to provide a total ordering, "$\prec$", on variables such that a variable's parents must be a subset of those variables less than it (i.e. $y \in \Pi_x$ only if $y \prec x$). We then ask the expert to indicate how strongly s/he believes each potential parent is a parent, measured in units of subjective probability. Denote this information by $E$. So for variables $x, y \in \mathcal{X}$ such that $y \prec x$, this is the prior probability that $y$ is a parent of $x$, denoted $Pr(y \rightarrow x \mid \prec, E)$. Assuming independence, a full prior on any given parent structure conditioned on the total ordering of variables is now

$$Pr(\Pi \mid \prec, E) = \prod_{x \in \mathcal{X}} Pr(\Pi_x \mid \prec, E) ,$$

assuming $\Pi$ is consistent with $\prec$, where

$$Pr(\Pi_x \mid \prec, E) = \left( \prod_{y \in \Pi_x} Pr(y \rightarrow x \mid \prec, E) \right) \cdot \left( \prod_{y \notin \Pi_x} (1 - Pr(y \rightarrow x \mid \prec, E)) \right) .$$

To extend this simple model of a partial theory we could also introduce correlations between potential parents.

So a partial Bayesian network is specified by a total ordering of variables $\prec$ together with a prior probability for each potential parent $E$, which allows us to evaluate $Pr(y \to x \mid \prec, E)$. To complete the prior, we need to specify $Pr(\theta \mid \Pi, \prec, E)$.

We assume $\theta$ is independent of $\prec$ and $E$ given $\Pi$ so develop a prior for $Pr(\theta \mid \Pi)$. We choose a prior that is a conjugate prior (it yields a posterior in the same functional form, so makes the mathematics simple (Berger, 1985)) and assumes the least amount of information is known about the conditional probability tables. This is a product of standard non-informative priors on multinomial distributions (each conditional probability distribution is a multinomial), the symmetric Dirichlet prior (Buntine, 1990b; Berger, 1985), and assumes prior independence between cells in the conditional probability table:

$$Pr(\theta \mid \Pi) = \prod_{x \in \mathcal{X}} \prod_{j \in v(\Pi_x)} \frac{\prod_{i \in v(x)} \theta_{i|j}^{\alpha_x - 1}}{Beta_{m_x}(\alpha_x, \ldots, \alpha_x)} ,$$

where $Beta_{m_x}$ is the $m_x$ dimensional Beta function given by

$$Beta_C(n_1, \ldots, n_C) = \frac{\prod_{i=1 \ldots C} \Gamma(n_i)}{\Gamma(\sum_{i=1 \ldots C} n_i)} ,$$

$\Gamma$ is the Gamma function, e.g. $\Gamma(n+1) = n!$, and $\alpha_x$ is a parameter to the prior for each variable $x$. A particular Bayesian network is often equivalent to a set of other Bayesian networks with some arc directions changed (Verma and Pearl, 1990). With

$$\alpha_x = \frac{\alpha}{m_x |v(\Pi_x)|} , \qquad (1)$$

this prior gives equivalent networks equivalent priors, and means marginal priors for individual variables are non-informative. (The proof of this is more involved than we have space for.)

## 4 Representing alternative Bayesian networks

Given a total ordering on variables, the theory refinement algorithm given in the next section considers reasonable alternative parent sets for each variable determined according to some criteria of reasonableness. For the variable $x$ alternative parent sets $\Pi_x$ will be a collection of subsets of $\{y : y \prec x\}$. Combining these gives a space of alternative parent structures that can then be represented by taking the cartesian product across $\mathcal{X}$ of the sets of reasonable parent sets. For each possible parent structure $\Pi$, we also have to know its posterior probability and sufficient information to update this given new examples. This space of parent structures and the additional information can be thought of as similar to a version space (Mitchell, 1982). However, because of the inherent uncertainty of the theories considered here, the "version space" cannot be updated by considering consistency with the training sample, most specific generalizations, etc. Instead Bayes theorem indicates the normative way of updating the "version space" of alternative parent structures and our posterior belief in them.

Unfortunately, the full space of parent structures is super exponential, so we cannot store and update details about each one. To overcome this we can store those whose posterior is quite high in relative terms since these are the only structures that are significant. This section outlines how we can calculate the posterior for a given parent structure, and how a reasonable set of alternative parent structures can be stored. We refer to this representation of the set of reasonable parent structures, their conditional probability tables and associated statistics as a *combined Bayesian network*. "Reasonable" in this context is given a more precise meaning in the next section where it is shown how to maintain and update combined Bayesian networks.

Let $P_x$ denote a set containing sets of reasonable parent variables for the variable $x$, so we have fair belief that the "true" $\Pi_x \in P_x$. Then the space of reasonable parent structures $\Pi$ given the total ordering $\prec$ is given by the cartesian product $\bigotimes_{x \in X} P_x$. Let the number of different reasonable parents $\sum_{x \in \mathcal{X}} |P_x|$ be denoted by $P$ (we note this now because it is useful in determining the operation count for later algorithms).

Each reasonable parent structure $\Pi$ has an associated subjective posterior probability indicating how strongly we currently believe it is the "true" structure. Having seen the sample *Sample*, and obtained the information $\prec$ and $E$ from the expert, this is $Pr(\Pi \mid Sample, \prec, E)$. According to standard rules of probability, this can be calculated as

$Pr(\Pi \mid Sample, \prec, E)$

$\propto Pr(\Pi \mid \prec, E) \cdot \int_\theta Pr(\theta \mid \Pi) Pr(Sample \mid \Pi, \theta)$

$$= \prod_{x \in \mathcal{X}} Pr(\Pi_x \mid Sample, \prec, E) \qquad (2)$$

where

$Pr(\Pi_x \mid Sample, \prec, E)$

$\propto Pr(\Pi_x \mid \prec, E)$

$$\prod_{j \in v(\Pi_x)} \frac{Beta_{m_x}(n_{x=1|j} + \alpha_x, \ldots, n_{x=m_x|j} + \alpha_x)}{Beta_{m_x}(\alpha_x, \ldots, \alpha_x)} ,$$

and $n_{x=i|j}$ is the number of examples in the training sample *Sample* with $x = i$ and $\Pi_x = j$, assuming every example in *Sample* has variable values fully specified. The solution to the integral follows by using standard properties of the Dirichlet integral (Buntine, 1990b). The counts $n_{x=i|j}$ are the only parameters in the posterior affected by the training sample and they are referred to as sufficient statistics (Berger, 1985); these need to be maintained during incremental learning.

Finally, each reasonable parent structure also has estimates for the parameters $\theta$ specifying the conditional probability tables. The estimated table for the variable $x$ is given by $\mathbf{E}_{\theta|Sample,\Pi}\left(\theta_{x=i|j}\right)$. According to standard rules of probability, these can be calculated as

$$\mathbf{E}_{\theta|Sample,\Pi}\left(\theta_{x=i|j}\right)$$
$$= \frac{\int_\theta \theta_{x=i|j} Pr(Sample \mid \Pi, \theta) Pr(\theta \mid \Pi)}{\int_\theta Pr(Sample \mid \Pi, \theta) Pr(\theta \mid \Pi)}$$
$$= \frac{n_{x=i|j} + \alpha_x}{n_{x=.|j} + m_x \alpha_x} \,, \qquad (3)$$

where $n_{x=.|j} = \sum_{i=1...m_x} n_{x=i|j}$. The integrations are done using standard properties of the Dirichlet integral and simplified using recursive properties of the Gamma function ($\Gamma(x+1) = x\Gamma(x)$).

With this basic information, we are now ready to describe the representation for a combined Bayesian network. In order to reconstruct the necessary conditional probability tables, compute the posteriors, etc. for each set of parent variables $\Pi_x \in P_x$, it is sufficient that the corresponding counts $n_{x=i|j}$ are kept. To save computation the posterior $Pr(\Pi_x \mid Sample, \prec, E)$ and the totals $n_{x=.|j}$ are also kept. To access all alternative parent sets $\Pi_x \in P_x$ efficiently they are stored in a lattice structure where subset and superset parent sets are linked together in a web, denoted the parent lattice for $x$. The full set of lattices is of size $P$ which is $\geq |\mathcal{X}|$. Because this is potentially exponential in $|\mathcal{X}|$, only those parent sets with significant posterior probabilities are stored and linked. For instance, we might only store those parent sets with posterior within a factor of 1/1000 of the maximum posterior parent set found so far to, for instance, restrain $P$ to be $O(|\mathcal{X}|)$. The structure updating algorithm does this. By increasing this factor close to 1, we are always guaranteed to make the full set of lattices manageable in size but at the expense of losing accuracy in theory refinement. But because posterior probabilities usually vary exponentially in learning, the set of reasonable parent sets should be manageable.

The root node of the parent lattice for $x$ is the empty set and the *leaves* are the sets $\Pi_x$ which have no supersets contained in $P_x$. We refer to this entire representation as a combined Bayesian network. Notice that we can easily fill in a lattice $P_x = \{\{a\}, \{a,b\}, \{a,c\}, \{a,d\}\}$ by adding $\{a,b,c\}$ and $\{a,c,d\}$ or $\{a,b,c,d\}$ to reduce the number of leaves, although some of these new leaves may have insignificant posterior probabilities.

To assist in the search and update of the lattice during theory refinement, nodes (i.e. parent sets and associated statistics) are labeled as alive, dead or asleep. Alive nodes represent the set of "reasonable" alternatives having high posteriors, and correspond to those parent sets in $P_x$. Dead nodes exist in the lattice as dead-end markers in the search space, they have been explored, have been forever determined as "unreasonable" alternatives and are not to be further expanded. Asleep nodes are similar but are only considered unreasonable for now and may be made alive later on. Furthermore, nodes can be either open or closed, depending on whether they require further expansion during search.

## 5   Theory Refinement

This section proposes several algorithms for the modification and interrogation of a combined Bayesian network. Most algorithms are linear-time in $|v(\Pi_x)|$, $|\mathcal{X}|$, $P$, which itself may be $O(|\mathcal{X}|)$, and other relevant variables. The structure update algorithm is an adjustable search algorithm so its time can vary from anything to fast greedy search to a slower beam search.

### 5.1   Parameter Updates

When the training sample *Sample* is extended, and we require a rapid incremental update of the combined Bayesian network, then a simple parameter update can be done without altering the structure of the parent lattices. This means, for each variable $x \in X$ and for each reasonable parent set $\Pi_x \in P_x$, we have to increment the corresponding cell counts, and update the posteriors. Normally, this process should effect only the alive nodes in the parent lattice. For instance, suppose *Sample* is extended to *Sample'* with the new example having $x = i$ and $\Pi_x = j$, then we should increment $n_{x=i|j}$ and

$$Pr(\Pi_x \mid Sample', \prec, E)$$
$$= Pr(\Pi_x \mid Sample, \prec, E)$$
$$\frac{(n_{x=i|j} + \alpha_x)(n_{x=.|j} - 1 + m_x \alpha_x)}{(n_{x=.|j} + m_x \alpha_x)(n_{x=i|j} - 1 + \alpha_x)} \,.$$

This follows from recursive properties of the Gamma function. The full update process will therefore take $O(P)$ operations. If we increase *Sample* by adding $N$ extra examples in a batch then we can repeat this process $N$ times. This process can be further sped up by initially updating only the leaf nodes in the parent lattices because the change in example counts can then be filtered upwards without reference to the examples.

### 5.2   Structure Updates

Given additional time, an any-time search can be begun to extend and modify the reasonable parent structures $P_x$ and the corresponding parent lattices to ensure high posterior parent sets are represented. This algorithm is first presented here as a one-time batch algorithm (starting from an empty lattice), and then differentiated to produce the incremental version. The algorithm presented is a simple beam search algorithm with three parameters such that $1 > C > D > E$. These are used to vary the search, as explained below. Alternatively, a branch and bound algorithm could be developed using upper bounds on posterior probabilities, or a corresponding decision theoretic search algorithm.

The batch beam search algorithm finds many parent sets with posteriors within a given factor $C$ of the best found. The beams searched are those parents sets within a factor $D$ of the best found. The algorithm is presented in pseudo-code in Figure 2. This search

---

**Input:** A variable $x$ and a prior on its parent sets $Pr(\Pi_x)$, and a training sample.

**Output:** The parent lattice for $x$ corresponding to this sample.

**Algorithm:** Set *Best-posterior* to the posterior for $\Pi_x = \emptyset$. Set *Open-list* to $\{\emptyset\}$. This maintains a list of parent sets within a factor $C$ of *Best-posterior*, those to be further expanded during search. Set *Alive-list* to $\{\emptyset\}$. This maintains a list of parent sets within a factor $D$ of *Best-posterior*, that are considered alive in the parent lattice. Repeat the process below until *Open-list* becomes empty. Take the top parent set $\Pi_x$ from *Open-list*. If its posterior is $< E \cdot \textit{Best-Posterior}$, mark this parent set dead. If its posterior is $< D \cdot \textit{Best-Posterior}$ ignore this parent set and proceed. Otherwise, generate all its children and calculate their posterior probabilities conditioned on the training sample. If the greatest posterior is $> \textit{Best-Posterior}$, then update *Best-Posterior* and modify *Alive-list* to reflect the new maximum. Mark all children with posterior $< E \cdot \textit{Best-Posterior}$ such that the sample size is $O(|\mathcal{X}||v(\Pi_x)|)$ as dead. Add all children with posterior $> D \cdot \textit{Best-Posterior}$ to *Open-list*. Add all children with posterior $> C \cdot \textit{Best-Posterior}$ to *Alive-list* and mark them as alive. Mark all remaining unmarked children as asleep.

---

Figure 2: The batch learning algorithm

is made easier by the fact the posterior probabilities on alternative structures tend to vary exponentially as structures change, and high posterior structures tend to clump together. This makes the beam search more efficient. Also, parent sets are marked dead if at any time they have a posterior less than a factor $E$ of the best and have fairly stable probability estimates. Many parent sets will be marked dead as posteriors for poor parent structures decrease exponentially with increasing sample size. Since dead nodes cannot be expanded, this further reduces the search. Finally, notice that if $C$ and $D$ are set close to 1, then the algorithm becomes a greedy search for a high posterior parent set.

A process reproducing the result of this algorithm can be run incrementally. This would be needed when an additional batch of examples is received. If asleep nodes have not been updated with previous additional samples because the parameter update process of the previous section was used, then these asleep nodes should first have their parameters updated and *Best-Posterior* recalculated. Processing after this is interruptible to achieve the any-time feature of the search. Adjust *Alive-list* and *Open-list* to reflect the new *Best-Posterior*. Finally expand nodes from *Open-list* and continue with the search. Some nodes may oscillate on and off *Alive-list* and *Open-list* because the posterior ordering of parent sets will oscillate as the training samples increases and the posteriors are modified. This is the problem of repeated restructuring reported by Crawford to occur in incremental learning algorithms (Crawford, 1989). This can be prevented by making a differential on $C$ and $D$ between placing a node on and taking a node off.

### 5.3 Structure posteriors

One useful form of feedback to the expert is to return information in exactly the same format initially obtained from the expert, a partial Bayesian network. This means calculating the posterior probability (conditioned on the training sample) that variable $y$ will be a parent of $x$

$$Pr(y \dashrightarrow x \mid Sample, \prec, E) = \sum_{\Pi_x \in P_x \,\wedge\, y \in \Pi_x} Pr(\Pi_x \mid Sample, \prec, E) .$$

The full calculation for all variables will take $O(|\mathcal{X}| \cdot P)$ operations. This information could be pictorially represented as a graph with arcs done in shades of grey to indicate strength of belief. Standard asymptotic properties of Bayesian methods assure us that as the sample size gets arbitrarily large, these posterior probabilities will converge to either 0 or 1.

### 5.4 Alternative Bayesian networks

Another useful form of feedback for the expert is to return some "good" Bayesian networks stored in the combined Bayesian network. We can do this by selecting for each $x \in \mathcal{X}$, a set of parents $\Pi_x$ and an associated conditional probability distribution. To ensure these are truly representative networks, we can return a collection of networks together in a compressed format corresponding to a single Bayesian network, denoted a *smoothed Bayesian network*. A similar operation has been presented for class probability trees (Buntine, 1990a).

For each variable $x$, we choose a leaf $L_x \in P_x$ from the parent lattice for $x$ using a probabilistic method described later. This provides one potential parent set for $x$. However, there may be more high posterior parents sets in $P_x$ that are subsets of $L_x$. We shall average each of their corresponding conditional probability tables together to obtain a single representative conditional probability table.

Denote by $S_x$ the set of parent sets that are subsets of $L_x$,

$$S_x \;=\; \{\, \Pi_x \,:\, \Pi_x \in P_x \wedge \Pi_x \subseteq L_x \,\} \;\subseteq\; P_x \;.$$

Then we can merge all these parent sets and average their conditional probability tables together to obtain a single representation of them all. This is done with the following formulae: the posterior probability that the "true" set of parents for $x$ is in $S_x$,

$$Pr(S_x \mid Sample, \prec, E) \;=\; \sum_{\Pi_x \in S_x} Pr(\Pi_x \mid Sample, \prec, E) \,,$$

the posterior expected conditional probability table for $x$ conditioned on $L_x$ assuming that the "true" set of parents for $x$ is in $S_x$,

$$E_{\Pi_x,\theta \mid S_x, Sample, \prec, E} \left( Pr(x = i \mid L_x = j, \Pi_x, \theta) \right)$$
$$= \sum_{\Pi_x \in S_x} Pr(x = i \mid \Pi_x = j_{\mid \Pi_x}, \Pi_x, Sample)$$
$$\cdot \frac{Pr(\Pi_x \mid Sample, \prec, E)}{Pr(S_x \mid Sample, \prec, E)} \,,$$

(note the 1st probability on the right-hand side of the equation is calculated using Equation (3)) and the posterior expected probability that $y$ is a parent of $x$ assuming that the "true" set of parents for $x$ is in $S_x$,

$$Pr(y \to x \mid S_x, Sample, \prec, E)$$
$$= \frac{\sum_{\Pi_x \in S_x \,\wedge\, y \in \Pi_x} Pr(\Pi_x \mid Sample, \prec, E)}{Pr(S_x \mid Sample, \prec, E)} \,.$$

We use these formulae as follows: for each $x$ we choose a leaf $L_x$ in $P_x$ randomly in proportion with $Pr(S_x \mid Sample, \prec, E)$. For the full set of variables this takes $O(P)$ operations. Because this process relies on selection of leaves from the parent lattice, it may be advantageous to reduce the number of leaves, as discussed with the structure update algorithm. For a variable $x$, we can display its set of parents pictorially using grey scales as discussed previously, but using $Pr(y \to x \mid S_x, Sample, \prec, E)$ as the probability $y$ is a parent of $x$. For the full set of variables this takes $O(|\mathcal{X}| \cdot P)$ operations. Finally, we can generate the conditional probability tables for $x$ given the value of $L_x$ by computing $E_{\Pi_x,\theta \mid S_x, Sample, \prec, E} \left( Pr(x = i \mid L_x = j, \Pi_x, \theta) \right)$. This represents the average of the various conditional probability tables corresponding to parent sets in $S_x$. Empirically, this has the effect of smoothing the conditional probability tables for $x$ given $L_x$ computed using Equation (3). This takes $O(\sum_{x \in \mathcal{X}} m_x |v(L_x)||S_x|)$ operations.

Given only a small training sample, this technique is likely to produce many different smoothed Bayesian networks corresponding to the many different alive leaves in the parent lattices. Perusal of these will give the expert some idea of the current variability in choice of a "good" Bayesian network. As the training sample size increases, asymptotic properties of Bayesian methods assure us the sets of high posterior parents and their conditional probability tables will become roughly equivalent so the different smoothed Bayesian networks produced will differ much less and eventually converge.

## 6 Extensions

This section briefly considers relaxing one of the assumptions made in the previous section: full conditional joint distributions exist at each node. Further extensions would be the handling of "missing values", where some examples have variable values missing, and the handling of expert designated "hidden variables" in the structure. Both problems can be handled the EM algorithm (Dempster et al., 1977).

While full conditional joint distributions are more general than any other model, their specification requires an exponential number of parameters. When estimating parameter values from data, this can be a severe problem as it is when trying to elicit the same probabilities from an expert. One way around this is to introduce approximate distributions of lower dimension. We have two issues to consider here: (1) How do you learn parameters for a specific conditional distribution? (2) How do you then patch the distribution learnt into the broad framework given previously?

There are many ways of representing restricted conditional probability distributions: trees (Buntine, 1990a), logistic regression and other qualitative models popular in economic statistics (Amemiya, 1985), and the noisy-or gate popular in AI (Pearl, 1988).

The noisy-or gate is described as follows. Suppose boolean variable $x$ is conditioned on boolean variables $x_1, \ldots, x_n$. The noisy-or has parameters $q_0, \ldots, q_n$,

$$Pr(x \mid x_1, \ldots x_n, q)$$
$$= q_0 \prod_{i=1,\ldots n} q_i^{1_{x_i}} \qquad \text{when } x \text{ is false, and}$$
$$= 1 - q_0 \prod_{i=1,\ldots n} q_i^{1_{x_i}} \qquad \text{when } x \text{ is true,}$$

where the indicator function $1_{x_i}$ is 1 when $x_i$ is true and zero otherwise. A similar conditional probability distribution is the multivariate logistic regression function, which in a slightly modified form applies to boolean variable $x$ conditioned on to boolean variables $x_1, \ldots, x_n$ and is

$$Pr(x \mid x_1, \ldots x_n, r)$$
$$= \frac{r_0 \prod_{i=1,\ldots n} r_i^{1_{x_i}}}{1 + r_0 \prod_{i=1,\ldots n} r_i^{1_{x_i}}} \qquad \text{when } x \text{ is false, and}$$

$$= \frac{1}{1 + r_0 \prod_{i=1,...n} r_i^{1_{x_i}}} \quad \text{when } x \text{ is true.}$$

(It is usually given with parameters $r_i = e^{r'_i}$.) The two forms approximate each other when the product is small. More generally, the logistic function is a symmetrized version of the noisy-or. Versions of the function exist when the variables are many-valued discrete variables, and to introduce higher-order correlations between variables. The logistic function has the same functional form as a simple (or "idiot") Bayes classifier, and can be obtained by taking the conditional distribution from a quadratic exponential distribution on discrete variables $x, x_1, \ldots x_n$.

To incorporate these methods into the framework just given, we need to be able to calculate the posterior expected parameter values, and the (relative) posterior probability that the noisy-or function or the logistic regression function is "true", independently of the parameter values. Since each conditional distribution is associated with a particular set of parent variables, the parameter values and the posterior can then be placed in a parent lattice of the combined Bayesian network. The posterior can be used, for instance, to search the space of logistic regressions over different parent sets using the algorithm of Figure 2, and also used when determining structure posteriors.

Posterior expected parameter values, and the (relative) posterior probability for both noisy-or and logistic regression models are readily estimated using standard maximum likelihood and Bayesian methods. It is simple to show that the sample likelihood functions for both the noisy-or and the logistic regression function is convex. So with a dominant likelihood term, the posterior on the parameters is unimodal and the maximum posterior parameters can be found using search methods such as scoring, Newton-Raphson, or conjugate gradient (Amemiya, 1985). A multivariate normal approximation for the posterior at this point (Berger, 1985, p224) can then be used to marginalize out the parameters and approximate the posterior probability that the noisy-or function of the logistic regression function is "true". Notice that because the numeric search algorithms are iterative, they are readily placed in an incremental framework. Given a few new training cases, start the iterative search at the previous maximum posterior point and convergence will be rapid to the new point (because the posterior is unimodal, there will be no catastrophic changes of the maximum posterior point).

## 7  Conclusion

This paper has presented a representation and some theory refinement algorithms for learning Bayesian networks. These have the following important properties:

- The representation can be initiated with a partial Bayesian network that quantifies the expert's experience and confidence. A similar approach was suggested in (Srinivas et al., 1990). Thereafter the representation maintains several reasonable hypotheses in a form of version space.

- The algorithms approximate the normative Bayesian solution to the corresponding batch learning problem. An analogous approximation for class probability trees significantly outperformed standard statistical and AI methods (Buntine, 1990a) on a large range of problems. A weaker approximation for batch learning (which finds a single high posterior network) has been reported to work well empirically (Cooper and Herskovits, 1991), and the parameter updating component of the algorithm corresponds to previous work (Spiegelhalter and Lauritzen, 1989).

- There is an incremental algorithm that allows any-time return for varied processing times between receipt of new examples. The development of the algorithm illustrates how a batch learning algorithm can be converted to an incremental learning algorithm.

- There are several algorithms that allow a user to interrogate the current hypotheses about Bayesian networks and to get some idea of their variability.

- The algorithms have parameters that allow fuller approximation of the normative solution. These parameters allow one to trade-off space/time complexity with (average-case) quality of learned theories (compare with (Buntine, 1990a; Henrion, 1990)).

- Extensions have been suggested to show how to handle different conditional probability models such as noisy-or gates and logistic functions.

Experience with a similar approach for learning trees suggests the algorithms, with some additional hacking, should work well.

## Acknowledgements

Several of these ideas have been suggested independently by Bob Fung, and I have also benefited from discussion with him. The writing of this paper was motivated by some comments made by Pat Langley.